# A Cascaded Residual UNET for Fully Automated Segmentation of Prostate and Peripheral Zone in T2-weighted 3D Fast Spin Echo Images

Lavanya Umapathy[1,2], Wyatt Unger[2], Faryal Shareef[2], Hina Arif[2], Diego Martin[2], Maria Altbach[2], and Ali Bilgin[1,2,3]

[1]*Electrical and Computer Engineering, University of Arizona, Tucson, Arizona, United States,* [2]*Medical Imaging, University of Arizona, Tucson, Arizona, United States,* [3]*Biomedical Engineering, University of Arizona, Tucson, Arizona, United States*

## ABSTRACT

Multi-parametric MR images have been shown to be effective in the non-invasive diagnosis of prostate cancer. Automated segmentation of the prostate eliminates the need for manual annotation by a radiologist which is time consuming. This improves efficiency in the extraction of imaging features for the characterization of prostate tissues. In this work, we propose a fully automated cascaded deep learning architecture with residual blocks, Cascaded MRes-UNET, for segmentation of the prostate gland and the peripheral zone in one pass through the network. The network yields high Dice scores (0.91±.02), precision (0.91±.04), and recall scores (0.92±.03) in prostate segmentation compared to manual annotations by an experienced radiologist. The average difference in total prostate volume estimation is less than 5%.

## INTRODUCTION

Prostate Cancer (PCa) is one of the most common cancers worldwide with nearly 1.28 million cases in 2018 (https://www.who.int/news-room/fact-sheets/detail/cancer). It is one of the leading causes of cancer death among American men[1]. Multi-parametric MR imaging (MP-MRI) has been shown to be effective in non-invasive diagnosis and staging of clinically significant PCa[2]. Along with anatomical information from qualitative T2-weighted imaging, MP-MRI can utilize quantitative parameter maps such as T2, T1, and Apparent Diffusion Coefficient (ADC).

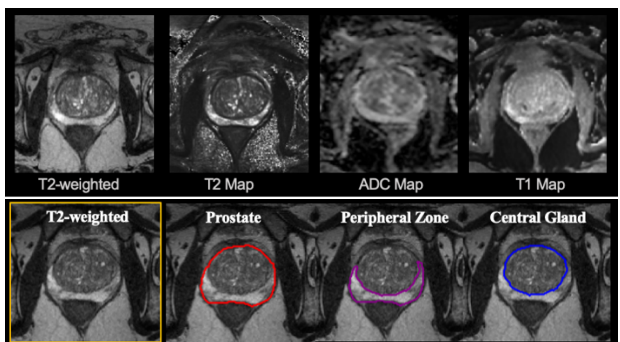

*Figure 1: Top row: Representative T2-weighted MR images of the prostate for a test subject. The corresponding quantitative maps (T2, ADC, and T1) for the image are also shown. Bottom row: Outlines for the prostate, peripheral zone, and central gland are shown for a different test subject.*

Figure 1 shows representative images for a subject from the MP-MRI protocol used in our work. The features obtained from these quantitative maps contain valuable information for the characterization of prostate tissue. The first step in automated MP-MRI processing is the segmentation of the prostate which eliminates the need for time-consuming manual annotations. Prostate segmentation can also provide total prostate volume (TPV) estimates to calculate volume-adjusted Prostate-specific antigen (PSA) values[3,4].

Deep learning networks, specifically, Convolutional Neural Networks (CNNs), have been used in a wide variety of medical image segmentation tasks, including prostate segmentation[5,6]. In this work, we propose a fully automated cascaded deep learning network architecture with residual blocks, Cascaded MRes-UNET, for the segmentation of the prostate gland and its sub-regions (central gland and peripheral zone) using isotropic resolution T2-weighted 3D Fast Spin Echo (FSE) images.

## MATERIALS AND METHODS

### Study Cohort and Annotations

3D T2-weighted FSE (SPACE) images with no fat saturation were acquired at 3T (Skyra, Siemens) on a total of 75 patients screened for prostate biopsy. The sequence parameters were as follows: in-plane resolution=1mm, slice thickness=1mm, matrix size=256x256, flip angle=120°, TR=1200ms, TE=238ms, average number of slices=112. Out of these 75 subjects, 67 subjects (~6000 images) were used for training and validation of the cascaded network. 8 subjects (748 images) were held out as test subjects to validate the generalizing ability of the proposed technique.

The prostate and central gland were manually annotated by an experienced radiologist on axial cross-sections of the T2-weighted images for each subject, using the sagittal cross-sections for reference. The ground truth for peripheral zone were obtained from the prostate and central gland masks and verified by the radiologist. Image pre-processing steps involved cropping the images to a smaller size (192x192) and signal normalization (zero mean and unit standard deviation). Data augmentation was performed using a combination of random in-plane rotations (-10°, 10°), translations (10 ,10 pixels), and image flips.

**CNN Architecture and Implementation**

Figure 2 shows the architecture of proposed Modified Residual UNET (MRes-UNET) used in this work. This architecture is a modified version of 2D UNET with residual blocks within the analysis and synthesis paths. These residual blocks use 1x1 convolutions along the identity paths[7,8]. Furthermore, instead of the feature concatenations used in UNET, the proposed architecture uses feature addition.

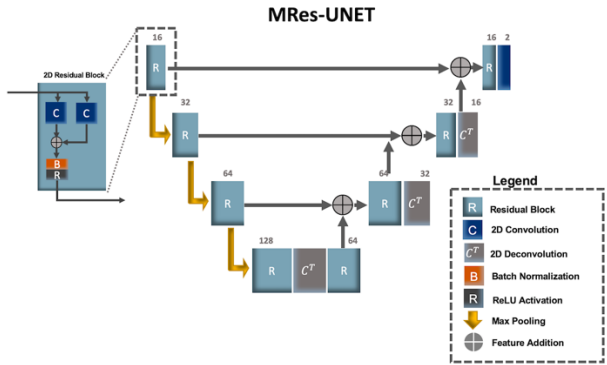

*Figure 3: An illustration of the Modified Residual UNET (MRes-UNET) architecture.*

The fully automated cascaded architecture consists of two sequential MRes-UNETs. Given an input T2-weighted image of the prostate, the first MRes-UNET predicts the mask for the prostate gland. The detected prostate mask is concatenated to the input image. The second MRes-UNET CNN uses this multi-channel data to predict the central gland within the prostate. The peripheral zone is identified using the central gland prediction as an exclusion mask within the prostate prediction.

To study the impact of the multi-channel input on the segmentation accuracy of the second MRes-UNET, we also trained a single-channel variant of the CNN. Here, we used the predicted prostate mask from the first MRes-UNET to extract the prostate region from within the image. This was then used as an input to the second MRes-UNET. We also trained a cascaded UNET[9] framework with a single-channel input to the second UNET, similar to Zhu et al[6]. We refer to this set-up as Cascaded UNET.

The CNNs were trained in Keras with Tensorflow[10] backend on an Ubuntu with Titan P100 GPU (NVIDIA). The following hyperparameters were used: loss=categorical cross entropy, optimizer=ADAM, learning rate=0.0005, batch size=5, epochs=50. The performance of the CNNs was evaluated using Dice score, precision, and recall. Bland Altman analysis was performed to analyze agreement between manual TPV and Cascaded MRes-UNET predicted TPV.

**RESULTS**

The end-to-end prediction time for a single T2-weighted image was 0.4 seconds. Figure 4 presents a qualitative evaluation of the proposed technique. The predictions are compared to manual annotations on two subjects, a normal subject with no PCa, and a second subject with low-grade cancer.

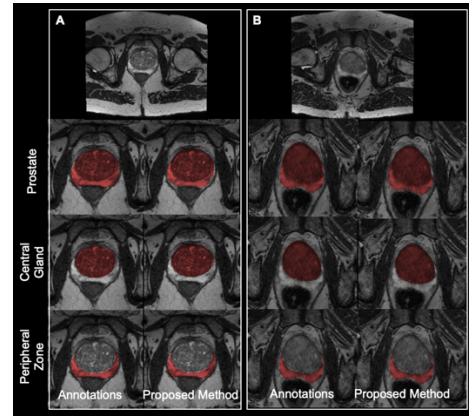

*Figure 2: Cascaded MRes-UNET predictions for (A) a normal subject with no prostate cancer (B) a subject with low grade cancer (Gleason score 6). The predictions for the prostate, central gland, and the peripheral zone agree well with manual annotations.*

Table 1 presents the average performance of the cascaded CNNs on the test set. For the cascaded MRes-UNET, we observed average Dice scores of 0.91, 0.90, and 0.73 for the prostate, central gland, and the peripheral zone, respectively. The cascaded UNET had comparable performance for the prostate but a poorer average Dice score for the peripheral zone (0.66 vs 0.73). Table 2 compares the difference in performance of the single-channel and the multi-channel variants of the MRes-UNET architecture on peripheral zone segmentation.

*Table 1: Comparison of the cascaded architectures on the test set*

|  |  | Dice | Precision | Recall |
|---|---|---|---|---|
| Cascaded MRes-UNET | Prostate | 0.91 ± 0.02 | 0.91 ± 0.04 | 0.92 ± 0.03 |
|  | Central Gland | 0.90 ± 0.03 | 0.89 ± 0.03 | 0.91 ± 0.04 |
|  | Peripheral Zone | 0.73 ± 0.11 | 0.79 ± 0.08 | 0.70 ± 0.13 |
| Cascaded UNET | Prostate | 0.91 ± 0.02 | 0.92 ± 0.02 | 0.91 ± 0.04 |
|  | Peripheral Zone | 0.66 ± 0.18 | 0.77 ± 0.11 | 0.59 ± 0.20 |

*Table 2: Comparison of single- and multi-channel variants of MRes-UNET for peripheral zone segmentation*

|  |  | Dice | Precision | Recall |
|---|---|---|---|---|
| MRes-UNET | Multi-Channel | **0.73** ± 0.11 | 0.79 ± 0.08 | 0.70 ± 0.13 |
|  | Single-Channel | **0.70** ± 0.14 | 0.76 ± 0.12 | 0.65 ± 0.17 |

The average difference in TPV between the ground truth and the predictions from Cascaded MRes-UNET was 4.79% ± 3.38%.

We observed a good correlation between TPV estimates from ground truth annotations and predictions (Figure 4). Bland-Altman analysis showed that all the data falls within the limits of agreement. The coefficient of variation was 6.2% and the reproducibility coefficient was 8.7%.

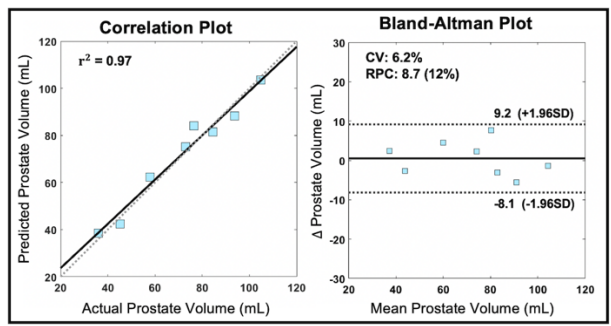

*Figure 4: Correlation plot (left) and Bland Altman analysis (right) for total prostate volume estimates from manual annotations and those from Cascaded MRes-UNET CNN. The values for coefficient of variation (CV) and reproducibility coefficient (RPC) are also presented.*

## DISCUSSION

An increase in prostate volume has been shown to be related to the rate of prostate cancer and as such accurate estimation of prostate volume is crucial. Prostate segmentation using the cascaded MRes-UNET framework yielded high average Dice scores (~0.91) with less than 5% error in TPV. In comparing the single and multi-channel variants of MRes-UNET, we observed that providing the prostate masks as an additional input to the second CNN in the cascade instead of extracting the prostate region helps in improving the average Dice scores, precision, and recall.

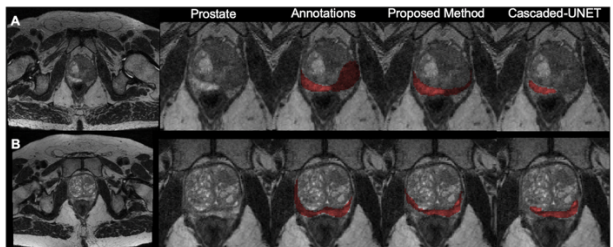

*Figure 5: Comparison of peripheral zone predictions from the CNNs for test subjects with (A) tumor in the left peripheral zone and an enlarged prostate and. (B) benign prostatic hyperplasia with enlarged central gland.*

Figure 5 shows how the segmentation accuracy of the CNNs is affected by abnormalities such as tumors in the peripheral zone or benign prostatic hyperplasia. These limitations can be overcome by using a larger training cohort that includes a multitude of conditions affecting the prostate sub-region contours.

## CONCLUSION

A cascaded fully automated MRes-UNET architecture was proposed for automated segmentation of prostate and peripheral zone from high-resolution T2-weighted FSE images. The segmentation masks obtained from the proposed approach can be applied to quantitative maps co-registered to the T2-weighted images to extract features of interest for the diagnosis and staging of prostate cancer.

### ACKNOWLEDGEMENTS

The authors would like to acknowledge support from the Technology and Research Initiative Fund (TRIF) Improving Health Initiative